\documentclass[10pt,twocolumn,letterpaper]{article}

\usepackage{iccv}
\usepackage{times}
\usepackage{epsfig}
\usepackage{graphicx}
\usepackage{amsmath}
\usepackage{amssymb}
\usepackage{subcaption}
\usepackage{multirow}
\usepackage{makecell}

\usepackage[breaklinks=true,bookmarks=false, colorlinks]{hyperref}

\iccvfinalcopy 

\def\iccvPaperID{****} 

\ificcvfinal\fi

\begin{document}

\title{SA-BEV: Generating Semantic-Aware Bird's-Eye-View Feature \\ for Multi-view 3D Object Detection}

\author{Jinqing Zhang\textsuperscript{\rm 1}, Yanan Zhang\textsuperscript{\rm 1}, Qingjie Liu\textsuperscript{\rm 1,2,3}\thanks{indicates the corresponding author.}, Yunhong Wang\textsuperscript{\rm 1,3}\\
\textsuperscript{\rm 1}State Key Laboratory of Virtual Reality Technology and Systems, Beihang University, Beijing, China\\
\textsuperscript{\rm 2}Zhongguancun Laboratory, Beijing, China\\
\textsuperscript{\rm 3}Hangzhou Innovation Institute, Beihang University, Hangzhou, China\\
{\tt\small \{zhangjinqing, zhangyanan, qingjie.liu, yhwang\}@buaa.edu.cn}
}

\maketitle
\ificcvfinal\thispagestyle{empty}\fi

\begin{abstract}
Recently, the pure camera-based Bird’s-Eye-View (BEV) perception provides a feasible solution for economical autonomous driving. However, the existing BEV-based multi-view 3D detectors generally transform all image features into BEV features, without considering the problem that the large proportion of background information may submerge the object information. In this paper, we propose Semantic-Aware BEV Pooling (SA-BEVPool), which can filter out background information according to the semantic segmentation of image features and transform image features into semantic-aware BEV features. Accordingly, we propose BEV-Paste, an effective data augmentation strategy that closely matches with semantic-aware BEV feature. In addition, we design a Multi-Scale Cross-Task (MSCT) head, which combines task-specific and cross-task information to predict depth distribution and semantic segmentation more accurately, further improving the quality of semantic-aware BEV feature. Finally, we integrate the above modules into a novel multi-view 3D object detection framework, namely SA-BEV. Experiments on nuScenes show that SA-BEV achieves state-of-the-art performance. Code has been available at https://github.com/mengtan00/SA-BEV.git.
\end{abstract}

\section{Introduction}

\begin{figure}[t]
    \centering
    \begin{minipage}[i]{2.7cm}
        \centering
        \includegraphics[width=2.7cm]{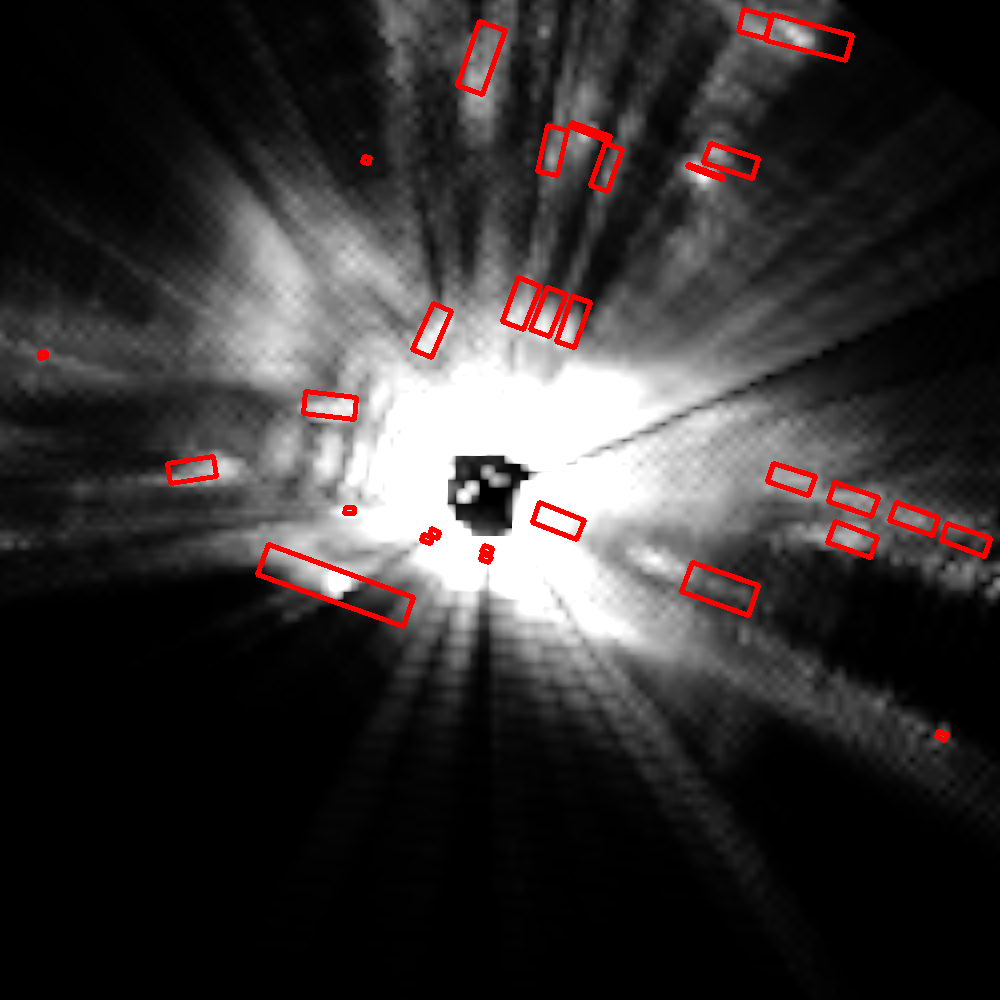}\\
        \includegraphics[width=2.7cm]{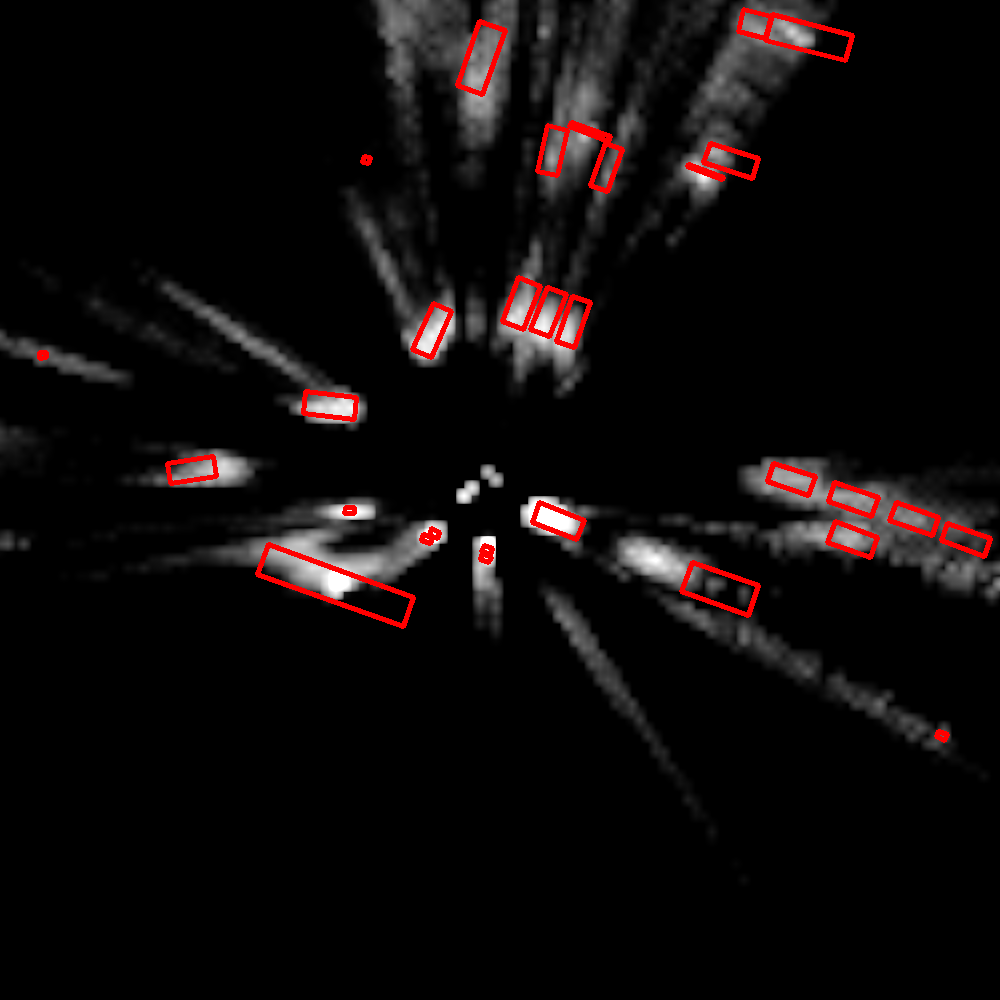}\\
        \subcaption*{Original Frame}
    \end{minipage}
    \begin{minipage}[i]{2.7cm}
        \centering
        \includegraphics[width=2.7cm]{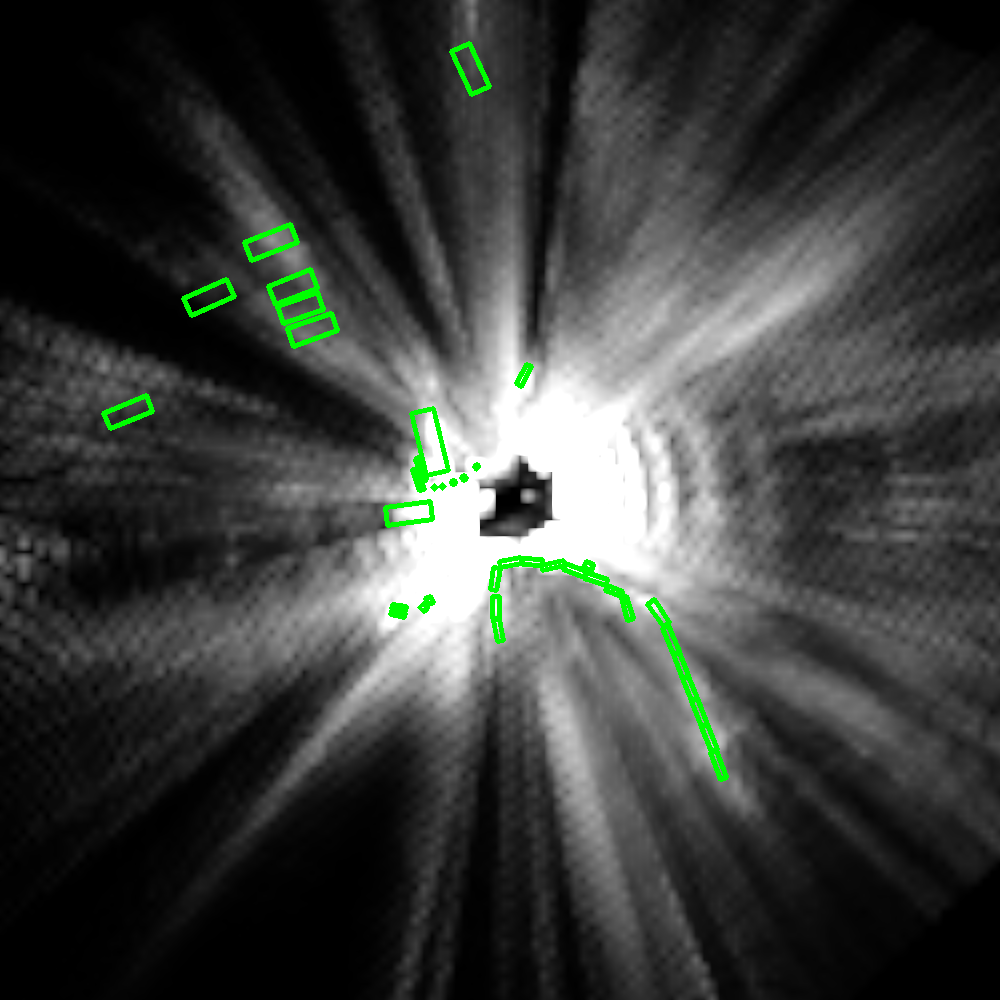}\\
        \includegraphics[width=2.7cm]{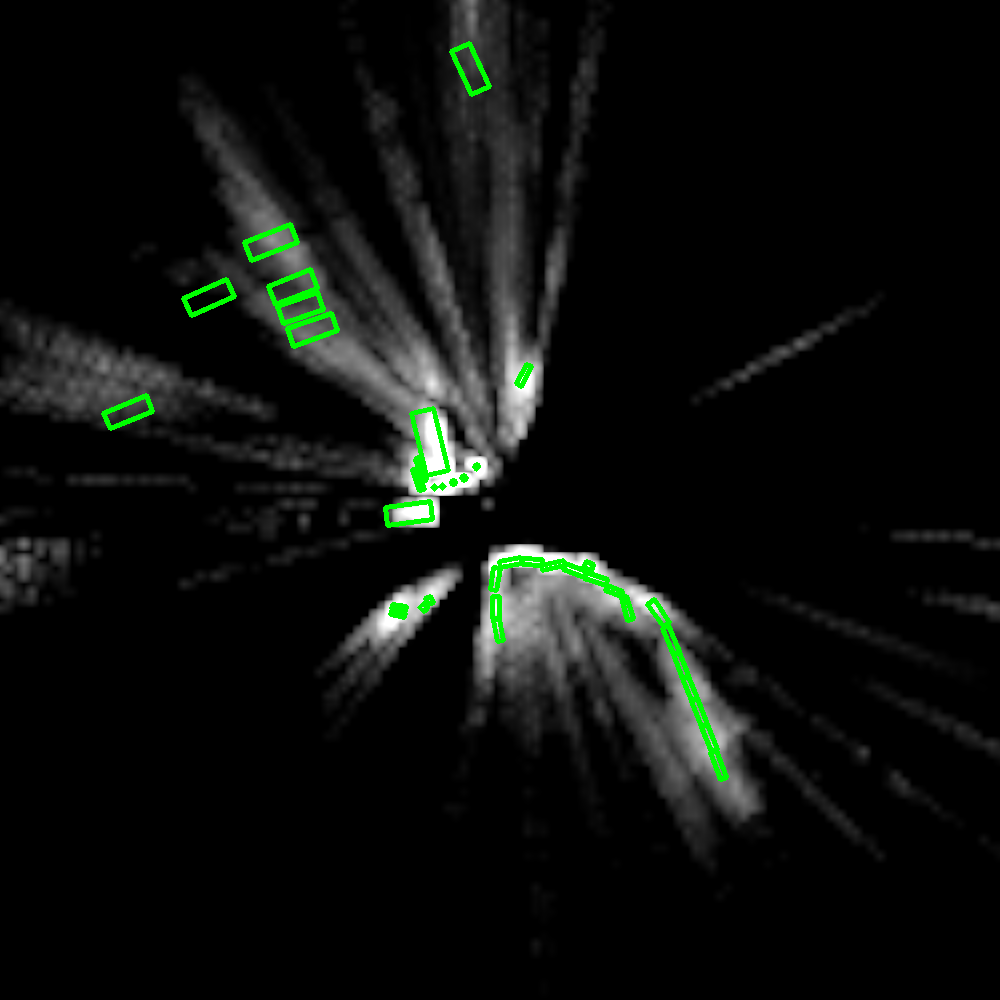}\\
        \subcaption*{Pasted Frame}
    \end{minipage}
    \begin{minipage}[i]{2.7cm}
        \centering
        \includegraphics[width=2.7cm]{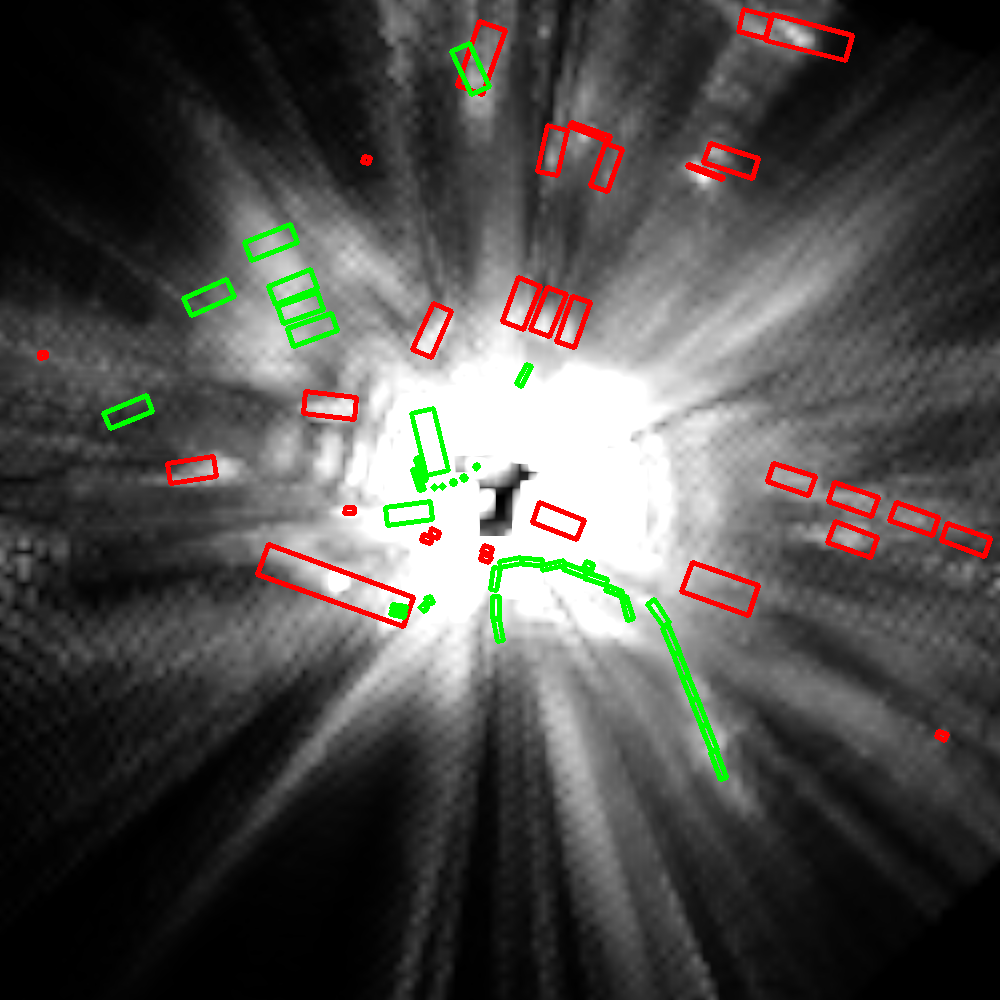}\\
        \includegraphics[width=2.7cm]{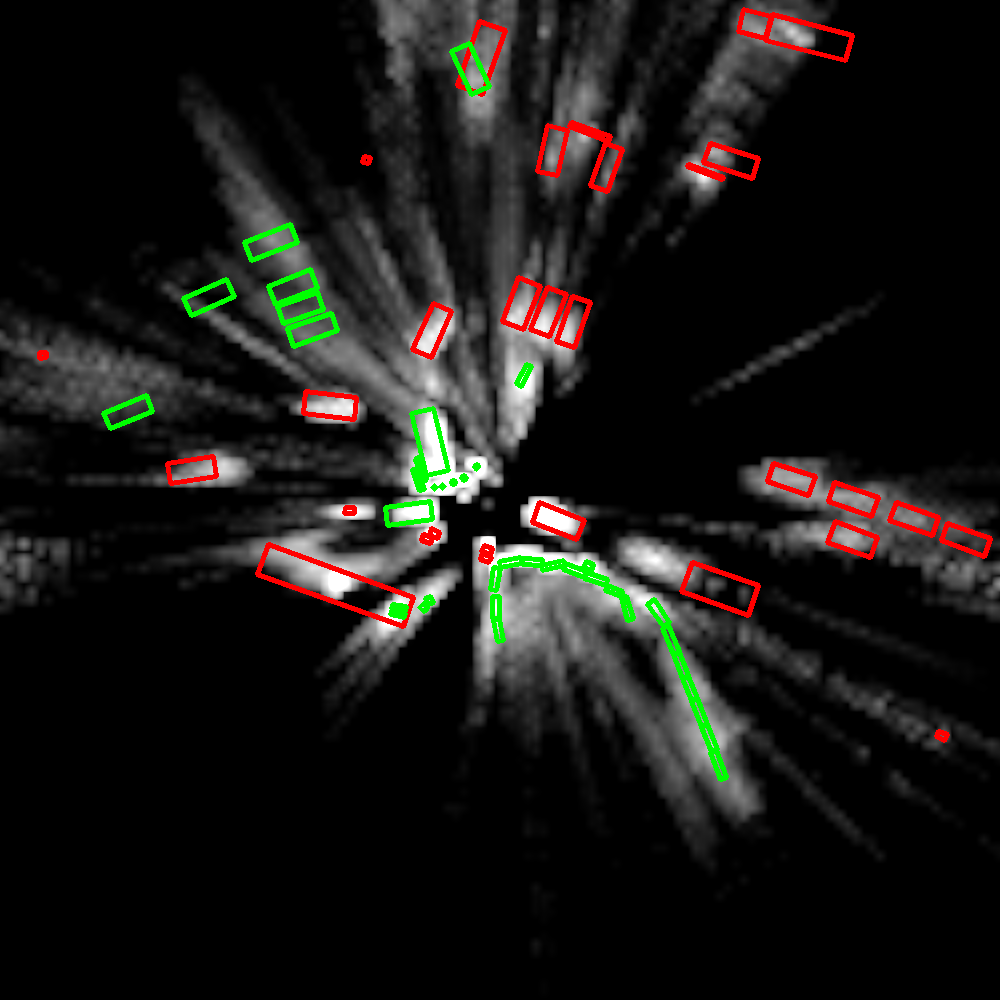}\\
        \subcaption*{After-pasting Frame }
    \end{minipage}
    \caption{Comparison between normal BEV features (upper row) and semantic-aware BEV features (lower row). The brightness reveals the norm of the features and the red / green boxes are the ground truth of the original / pasted frame. The last column shows BEV-Paste, an data augmentation strategy that matches semantic-aware BEV features.}
    \label{fig:bev}
\end{figure}

Camera and LiDAR are the two most commonly used sensors for 3D object detection, which is essential to autonomous driving systems. LiDAR-based methods~\cite{deng2021voxel, lang2019pointpillars, YanYan2018SECONDSE, zhang2021pc, yang20203dssd, yin2021center, zhou2023octr} have attained excellent performance due to the accurate spatial structure information of point clouds, but the expensive LiDAR sensor reduces its universality.
In contrast, camera-based methods~\cite{wang2021fcos3d, wang2022probabilistic, wang2022detr3d, liu2022petr, liu2022petrv2} are relatively low-cost with plentiful semantic information, but are constrained by the lack of geometric depth cues.

Considering the performance gap between camera and LiDAR, the Bird's-Eye-View paradigm transforms multi-view image features into the BEV feature to make the following 3D perception easier~\cite{philion2020lift, huang2021bevdet}. This practical and scalable camera-only paradigm is gaining popularity, and numerous advancements have allowed it to reach high perceptual precision~\cite{li2022bevdepth, li2022bevstereo, li2022bevformer, huang2022bevdet4d, jiang2022polarformer}. The core step of the BEV paradigm is generating virtual points from image features, which will be projected into the ``pillarized'' BEV space. The features of the virtual points in the same pillar are then cumulated as the BEV feature. However, this operation does not fully utilize the semantic information of the image features and will inject massive background information that submerges object information.

In order to take full advantage of the valuable semantic information of image features, we propose Semantic-Aware BEV Pooling (SA-BEVPool) to generate semantic-aware BEV features, which replace the normal BEV feature for 3D detection. Before projecting virtual points into BEV space, the semantic segmentation of image features is first predicted. If a virtual point is generated by the image element that belongs to the background, it will not be projected into BEV space. Similarly, virtual points with low depth scores will also be ignored. The comparison between normal BEV features and semantic-aware BEV features is shown in Fig. \ref{fig:bev}. SA-BEVPool can obviously filter out most of the background BEV features and alleviate the problem that the large proportion of background information submerges object information, therefore effectively improving the detection performance. Some multi-modal 3D objectors~\cite{vora2020pointpainting, yin2021multimodal} also adopt segmentation on images when combining with LiDAR features, but they generally use powerful instance segmentation networks like CenterNet2~\cite{zhou2021probablistic} to predict the segmentation of the large-scale image. Instead, SA-BEVPool can be easily applied in current BEV-based detectors like BEVDepth~\cite{li2022bevdepth} and BEVStereo~\cite{li2022bevstereo} by using their depth branch to simultaneously predict the semantic segmentation of small-scale image features.

GT-Paste~\cite{YanYan2018SECONDSE} is a successful data augmentation strategy that has been frequently adopted by various LiDAR-based 3D detectors. However, due to the modality gap, it cannot directly adapt to camera-based 3D detectors. In our work, thanks to the reliable depth distribution and semantic segmentation predicated on image features, the semantic-aware BEV feature can approximately represent the information of all objects that are located appropriately in BEV space. As a result, adding the semantic-aware BEV features of another frame to the current semantic-aware BEV feature is the same as pasting all objects of another frame into the current frame. This strategy, we called BEV-Paste, enhances data diversity in a similar way to GT-Paste.

Although it is convenient to predict depth distribution and semantic segmentation with the same branch, doing so may result in a sub-optimal semantic-aware BEV feature. Research conclusion in the field of multi-task learning demonstrates that the integration of specific tasks and cross-task information is more conducive to the optimal solution of multiple prediction tasks. Inspired by this, we design a Multi-Scale Cross-Task (MSCT) head to combine the task-specific and cross-task information through multi-task distillation and dual-supervision on multiple scales prediction.

We integrate our proposed modules as a whole and name it SA-BEV. Extensive experiments on nuScenes dataset show that SA-BEV achieves a new state-of-the-art. In summary, the major contributions of this paper are:
\begin{itemize}
    \item We propose SA-BEVPool, which uses semantic information to filter out unnecessary virtual points and generate the semantic-aware BEV feature, alleviating the problem that the large proportion of background information submerges the object information.
    \item We propose BEV-Paste, an effective and convenient data augmentation strategy closely matching the semantic-aware BEV feature, which enhances data diversity and further promotes detection performance.
    \item We propose the MSCT head that combines the task-specific and cross-task information through multi-task learning on multiple scales, facilitating the optimization of the semantic-aware BEV feature. 
\end{itemize}

\begin{figure*}
    \centering
    \includegraphics[width=14cm]{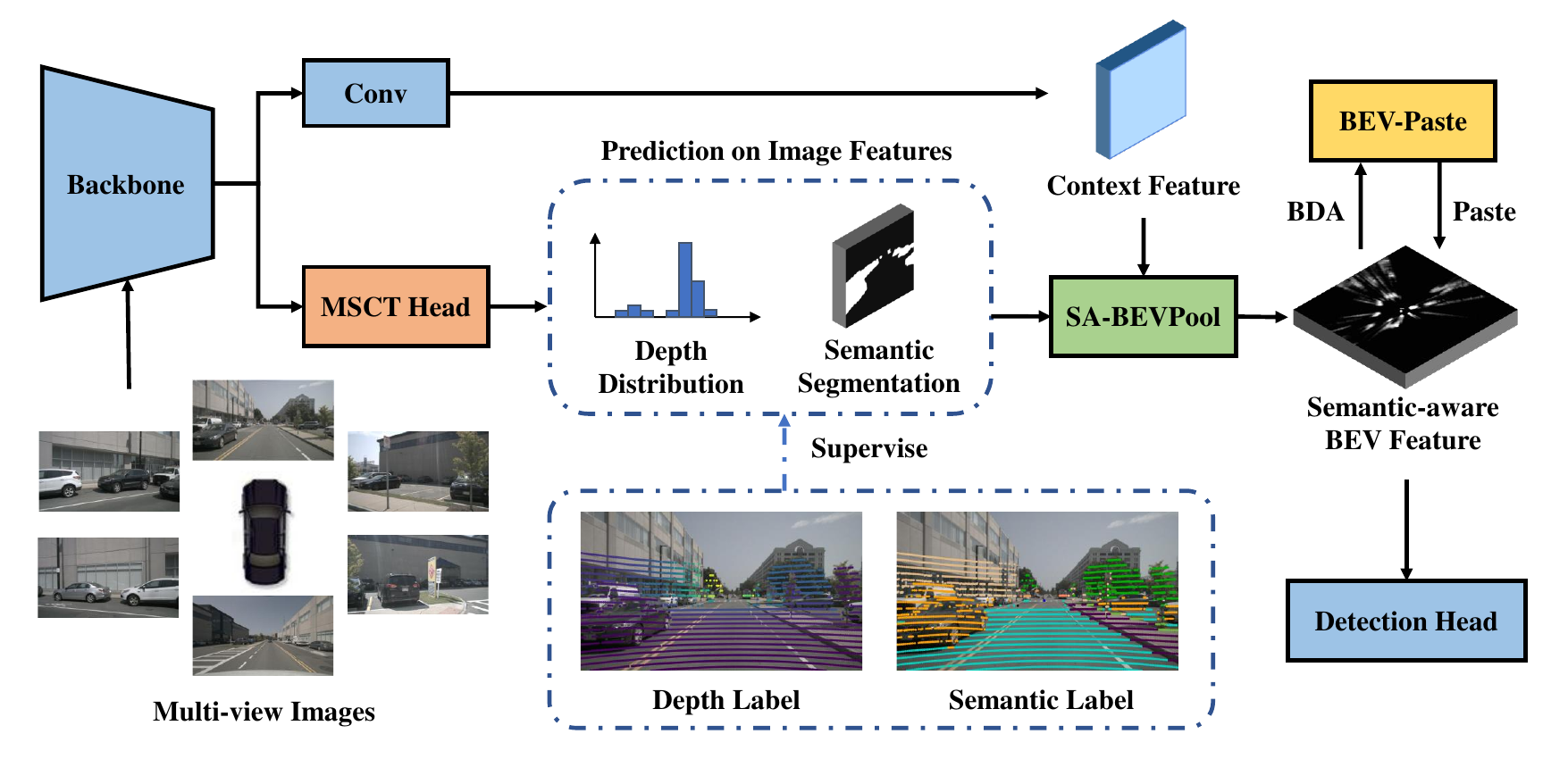}
    \caption{Overall framework of SA-BEV. The MSCT head uses multi-scale image features to predict the depth distribution and semantic segmentation, which are utilized by SA-BEVPool to generate the semantic-aware BEV feature. BEV-Paste is then applied to increase the diversity of BEV features during the training phase.}
    \label{fig:all_structure}
\end{figure*}

\section{Related Work}

\subsection{Vision-based 3D Object Detection}

Although camera does not provide reliable depth of the surroundings like LiDAR, the plentiful semantic information carried by images still supports vision-based 3D object detectors to achieve considerable precision. Early vision-based 3D detectors predict attributes of 3D objects directly from 2D image features. For instance, CenterNet~\cite{zhou2019objects}, a 2D detector, can be used to predict 3D objects without many modifications. Lately, FCOS3D~\cite{wang2021fcos3d} detects the 2D centers of the 3D objects, and features around the centers are used to predict the 3D attributes. PGD~\cite{wang2022probabilistic} establishes geometric relation graphs to improve the depth estimation results for better 3D object detection. DETR3D~\cite{wang2022detr3d} follows DETR~\cite{carion2020end} and detects 3D objects with Transformer. PETR~\cite{liu2022petr} introduces 3D position-aware representations, ameliorating the detection precision. PETRv2~\cite{liu2022petrv2} further brings in temporal information and improves efficiency.

Recently, an approach that transforms the image feature into the BEV feature is proposed by LSS~\cite{li2022bevdepth}, and BEVDet~\cite{huang2021bevdet} employs the detection head of CenterPoint~\cite{yin2021center} to predict 3D objects from BEV feature. This paradigm can achieve comparable accuracy without cumbersome operations and is easy to extend, making it gain popularity. BEVDet4D~\cite{huang2022bevdet4d} processes multiple key frames to introduce temporal information. BEVFormer~\cite{li2022bevformer} utilizes the deformable attention mechanism to generate the BEV feature. BEVDepth~\cite{li2022bevdepth} applies explicit depth supervision on the predicted latent depth distribution, improving the detection accuracy. BEVStereo~\cite{li2022bevstereo} further improves the quality of the depth by applying the multi-view stereo on nearby key frames. PolarFormer~\cite{jiang2022polarformer} generates the BEV feature using the polar coordinate for a more accurate location. However, these methods project all image features into BEV features, without considering the problem that the large proportion of background information may submerge the object information. In this paper, we propose SA-BEVPool, which can filter out background information according to the semantic segmentation of image features and generate semantic-aware BEV features.

\subsection{Data Augmentation in 3D Object Detection}

The diversity of the dataset is crucial to the generalization performance of models. Besides regular data augmentation like random scaling, flipping and rotation, GT-Paste~\cite{YanYan2018SECONDSE} is another effective strategy frequently used by LiDAR-based detectors. It crops the points according to the 3D boxes of ground truth and pastes them to other frames to create new training data. Lately, an improvement in generating a visibility map to correct the wrong occlusion relationship introduced by GT-Paste is proposed in~\cite{hu20wysiwyg}. The augmentation on the individual object is also proposed in~\cite{choi2021part, zheng2021se} which takes object points into parts and applies operations like dropout, swap, and mix on it.

Since GT-Paste shows excellent effectiveness on increase data diversity, there have been some attempts to adopt it in camera-only 3D detectors. Box-Mixup and Box-Cut-Paste proposed by~\cite{santhakumar2021exploring} directly cut the objects from images according to their 2D bounding boxes and paste them into other frames. To paste precisely, objects are cropped by their instance masks in~\cite{zhang2020exploring}. Pointaugmenting~\cite{wang2021pointaugmenting} utilizes a more complicated way to tackle the occlusion relationship of original objects and pasted objects. However, these attempts to expand the GT-Paste into image space cannot easily overcome the issues caused by the gap between LiDAR and camera. In this paper, we propose BEV-Paste, a convenient way to effectively extend GT-Paste into BEV-based methods with the help of SA-BEVPool. 

\subsection{Multi-task Learning}
 
Multi-task learning generally leads to better prediction through interactive learning between multiple tasks. According to~\cite{SimonVandenhende2021MultiTaskLF}, both task-specific information and cross-task information are important for getting optimal results on multiple tasks. PAD-Net~\cite{Xu_2018_CVPR} proposes multi-modal distillation module to automatically supplement cross-task information. PAP-Net~\cite{zhang2019pattern} extracts cross-task affinity patterns and recursively propagates the pattern by affinity matrices. MTI-Net~\cite{vandenhende2020mti} models the task interactions at different scales and aggregates multi-scale information to make precise predictions. 

Some methods also introduce multi-task learning into 3D object detection. MMF~\cite{Liang_2019_CVPR} deeply fuses the features of images and LiDAR through simultaneous supervision made on multiple tasks. Latent support surfaces are estimated in~\cite{ren20183d} to help improve the precision of 3D detection. A multi-task LiDAR network proposed by~\cite{feng2021simple} makes predictions on 3D detection and road understanding that can complement each other. Some BEV-based 3D detectors~\cite{li2022bevformer, liu2022petrv2} also apply BEV segmentation to obtain better BEV representation. In this paper, we propose the MSCT head that combines the task-specific and cross-task information of multiple scales for depth estimation and semantic segmentation, facilitating the optimization of the semantic-aware BEV feature.


\section{Method}

In this work, we propose SA-BEV, a novel multi-view 3D object detection framework that generates semantic-aware BEV features for better detection performance. It contains the Semantic-Aware BEV Pooling (SA-BEVPool), the BEV-Paste data augmentation strategy and the Multi-Scale Cross-Task (MSCT) head. The overall framework of SA-BEV is shown in Fig.~\ref{fig:all_structure}.

\subsection{Semantic-Aware BEV Pooling}
\begin{figure}
    \centering
    \includegraphics[width=7.5cm]{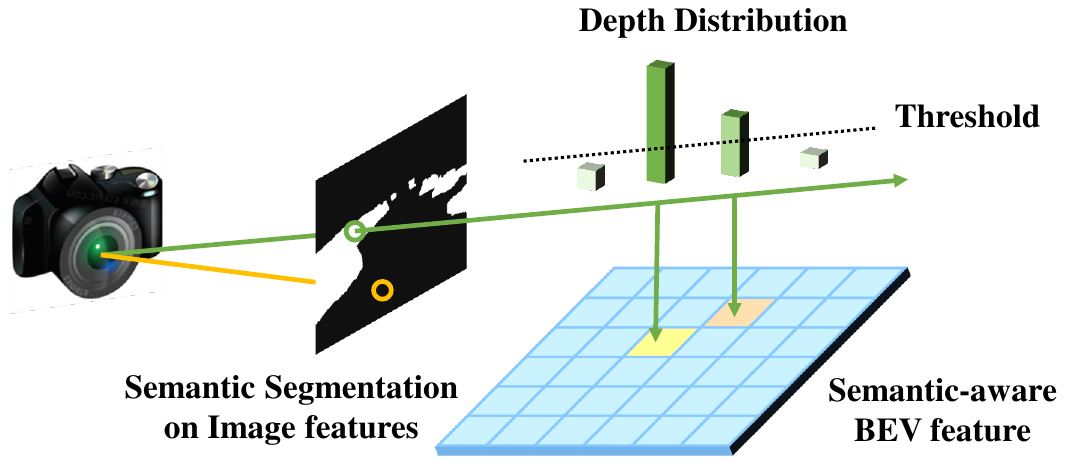}
    \caption{Illustration of Semantic-Aware BEV Pooling. The green line represents the projection of the foreground features, while the yellow line represents the ignored background features. The foreground virtual points with depth scores lower than the threshold are also ignored.}
    \label{fig:pool}
\end{figure}

The way of transforming the image features into BEV features for better perception was first proposed by LSS~\cite{philion2020lift}. It predicts the depth distribution $\boldsymbol{\alpha}$ and context feature $\boldsymbol{c}$ of each image feature element. Then each element generates virtual points at different depths. The feature of the point at depth $d$ is represented as $\boldsymbol{p}_d=\alpha_{d}\boldsymbol{c}$. After that, all virtual points will be projected to the BEV space which is divided into pillars. The features of virtual points in the same pillar will be cumulated as the BEV feature. This process is known as BEV pooling.

Subsequent BEV-based 3D detectors~\cite{li2022bevdepth, liu2022bevfusion, huang2022bevpoolv2} significantly improve the efficiency and accuracy of BEV pooling. But what is unchanged is that these methods insist on projecting all virtual points into BEV space. However, we argue that this is unnecessary for 3D detection tasks. On the contrary, if all virtual points belonging to the background are projected, the foreground virtual points that account for less than 2\% of the total virtual points will be submerged. It will confuse the following detection head and reduce the detection accuracy.

To highlight the valuable foreground information in the BEV features, we propose a novel Semantic-Aware BEV Pooling (SA-BEVPool) that is shown in Fig. \ref{fig:pool}. It applies semantic segmentation on the image features to get the foreground score $\beta$ of each element. The element with low $\beta$ is more possible to carry useless information for detection, and the virtual points generated from it will be ignored during the BEV pooling. Similarly, the virtual points with low $\alpha_d$ provide trivial information and will also be ignored. Denoting the filtering function as:
\begin{equation}
    \mathcal{F}(x,y)=
    \begin{cases}
    0, & x < y,\\ 
    1, & x \ge y
    \end{cases},
\end{equation}
the point features after filtering are changed as:
\begin{equation}
    \hat{\boldsymbol{p}}_d = \mathcal{F}(\alpha_d, T_D)\mathcal{F}(\beta, T_S)\boldsymbol{p}_d,
\end{equation}
where $T_D$ and $T_S$ are the threshold for $\alpha_d$ and $\beta$. Only non-empty $\hat{\boldsymbol{p}}_d$ will construct BEV feature. Since the operation of filtering relatively low-value virtual points attaches semantic information to the generated BEV features, they can be called semantic-aware BEV features.

The difference between the normal BEV feature and the semantic-aware BEV feature is clearly shown in Fig. \ref{fig:bev}. The normal BEV features in the first row generally have a ring of light in the center, which represents the ground. It accounts for most of the signal strength in the feature without contributing useful information for detection. In contrast, most of the background information is removed in the semantic-aware BEV feature and object information is emphasized. Furthermore, the location of object information in the semantic-aware BEV feature matches the ground truth well, making the detection head easier to predict accurately.

\begin{figure}
    \centering
    \includegraphics[width=7.5cm]{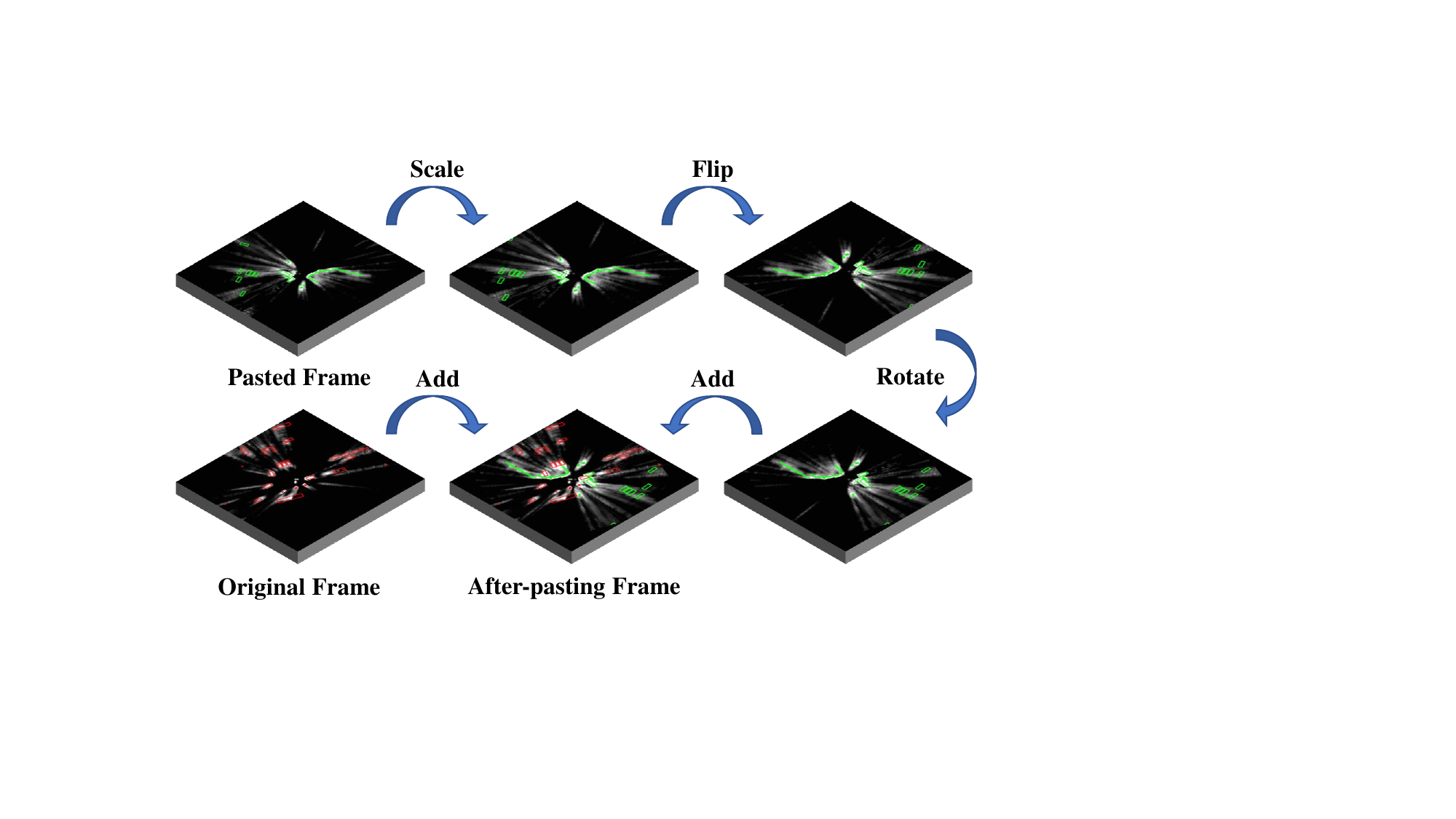}
    \caption{Illustration of BEV-Paste.}
    \label{fig:paste}
\end{figure}

\subsection{BEV-Paste}

GT-Paste~\cite{YanYan2018SECONDSE} is a data augmentation strategy commonly used by LiDAR-based 3D detectors. It has been proved that the diversity of the dataset can be effectively increased by sampling the points in 3D boxes and pasting them into other frames. However, several problems prevent the application of GT-Paste in camera-based methods. First, sampling an object by the bounding box on the image can not get its pure data as the point cloud does. Another problem is that pasting objects to another image may wrongly occlude original objects and result in data loss. In addition, the illumination change of different frames also gives the pasted objects unnatural appearances. Some multi-modal 3D detectors~\cite{santhakumar2021exploring, wang2021pointaugmenting, zhang2020exploring, zhang2022cat} make an effort to solve these issues but generally lack convenience and accuracy.

Here, we propose BEV-Paste that successfully applies GT-Paste in camera-only 3D detectors without complicated steps. With SA-BEVPool, the semantic-aware BEV features transformed from image features approximately represent the information of all objects in the frame as shown in Fig. \ref{fig:bev}. It makes adding arbitrary semantic-aware BEV features of two frames during the training phase equivalent to aggregating the objects contained in two frames into one frame. While effectively increasing the diversity of the entire training dataset, BEV-Paste does not increasing the computational cost in the inference stage.

In practice, we randomly select the original semantic-aware BEV feature $\bold{B}_O$ and the pasted semantic-aware BEV feature $\bold{B}_P$ from the same batch. This is to guarantee $\bold{B}_O$ and $\bold{B}_P$ follow the same distribution. Instead of directly pasting $\bold{B}_P$ to $\bold{B}_O$, extra BEV data augmentations (BDA) shown in Fig.~\ref{fig:paste} are first applied to $\bold{B}_P$ and $\hat{\bold{B}}_P$ is obtained. It prevents the data duplication of $\bold{B}_P$. The same augmentation is also applied to the ground truth of the pasted frame $G_P$ to get $\hat{G}_P$. The detection loss after BEV-Paste can be represented as:
\begin{equation}
    \mathcal{L}_{det}=\mathcal{L}_{det}(Det(\bold{B}_O+\hat{\bold{B}}_P),G_{O}\cup \hat{G}_P),
\end{equation}
where $Det$ includes BEV encoder and detection head, $G_{O}$ is the ground truth of original frame.

\subsection{Multi-Scale Cross-Task Head}

It is a convenient way to obtain semantic-aware BEV features by making the depth branch predict semantic segmentation at the same time, but it generally leads to sub-optimal results. Let us regard the generation of the semantic-aware BEV feature as a multi-task learning application. According to the research conclusion, both task-specific information and cross-task information are important for getting the global optimal solution of multiple tasks. If the depth distribution and semantic segmentation are predicted by the same network branch, the network only extracts cross-task information from the image features and can not perform optimally on each task.
\begin{figure}
    \centering
    \includegraphics[width=8.5cm]{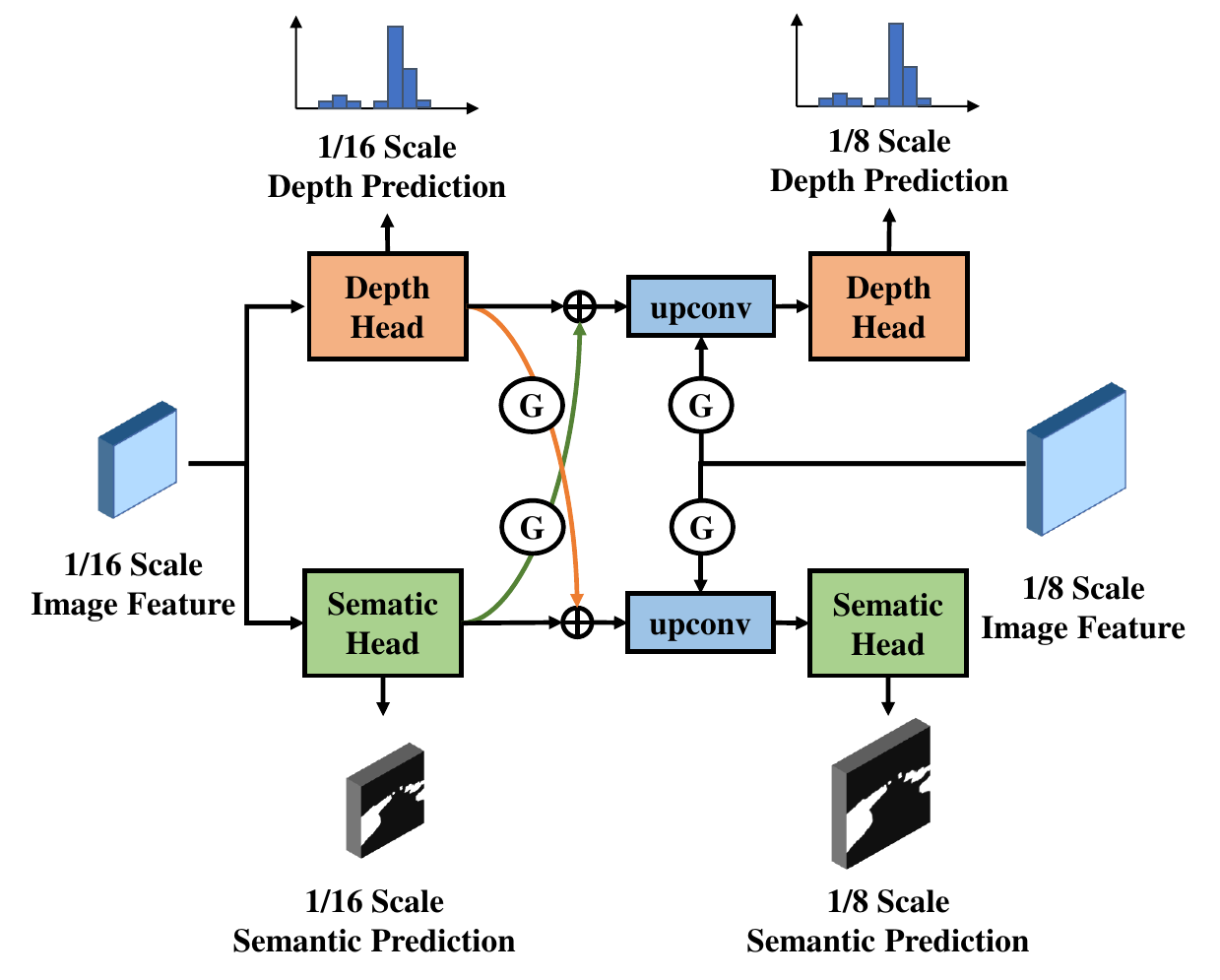}
    \caption{The structure of Multi-Scale Cross-Task head. Both 1/16 and 1/8 scale image features are taken as input.}
    \label{fig:mscthead}
\end{figure}

Inspired by the principle of multi-task learning, we design a Multi-Scale Cross-Task (MSCT) head as shown in Fig. \ref{fig:mscthead}. In the first stage, the head takes 1/16 scale image feature $\bold{F}_I^{16}$ as input and makes a relatively coarse prediction of depth distribution and semantic segmentation. After that, $\bold{F}_I^{16}$ is transformed into depth feature $\bold{F}_D^{16}$ and semantic feature $\bold{F}_S^{16}$, which carry the task-specific information of their own task. To complement cross-task information, Multi-Task Distillation (MTD) module proposed in~\cite{Xu_2018_CVPR} is applied between $\bold{F}_D^{16}$ and $\bold{F}_S^{16}$. It is composed of several self-attention blocks, which generate gate map $\mathcal{G}$ by
\begin{equation}
    \mathcal{G}(\bold{F})=\sigma(W_G\bold{F}),
\end{equation}
where $W_G$ is the gate convolution and $\sigma$ denotes sigmoid function.
The features supplemented by the cross-task information can be formulated as:
\begin{align}
    \hat{\bold{F}}_{D}^{16}&=\bold{F}_{D}^{16}+\mathcal{G}(\bold{F}_{D}^{16})\odot(W_t\bold{F}_{S}^{16}) \\
    \hat{\bold{F}}_{S}^{16}&=\bold{F}_{S}^{16}+\mathcal{G}(\bold{F}_{S}^{16})\odot(W_t\bold{F}_{D}^{16})
\end{align}
where $W_t$ is the task convolution and $\odot$ denotes element-wise multiplication. It is clear that MTD uses these self-attention blocks to automatically extract cross-task information from one task feature and add it to other task features.

After the task features interaction, $\hat{\bold{F}}_{D}^{16}$ and $\hat{\bold{F}}_{S}^{16}$ obtain both task-specific information and cross-task information. Before inputting them into the second stage prediction head, they are up-sampled to the 1/8 scale and combined with the 1/8 scale image feature $\bold{F}_{I}^{8}$ using the same self-attention blocks. The features can be formulated as:
\begin{align}
    \hat{\bold{F}}_{D}^{8}&=Up(\hat{\bold{F}}_{D}^{16})+\mathcal{G}(\bold{F}_{I}^{8})\odot(W_t\bold{F}_{I}^{8}) \\
    \hat{\bold{F}}_{S}^{8}&=Up(\hat{\bold{F}}_{S}^{16})+\mathcal{G}(\bold{F}_{I}^{8})\odot(W_t\bold{F}_{I}^{8})
\end{align}
The second stage head then predicts the relatively fine depth distribution and semantic segmentation which will be used to generate the semantic-aware BEV feature. 
\begin{table*}
    \begin{center}  
        \caption{Comparison with previous state-of-the-art multi-view 3D detectors on the nuScenes \textit{test} set.}
        \label{tab:test}
        \begin{tabular}{l|c|c|cc|ccccc}
            \hline
            Method&Backbone&Resolution&mAP$\uparrow$&NDS$\uparrow$&mATE$\downarrow$&mASE$\downarrow$&mAOE$\downarrow$&mAVE$\downarrow$&mAAE$\downarrow$\\
            \hline
            FCOS3D~\cite{wang2021fcos3d}&ResNet-101&900$\times$1600&0.358&0.428&0.690&0.249&0.452&1.434&0.124\\
            PGD~\cite{wang2022probabilistic}&ResNet-101&900$\times$1600&0.386&0.448&0.626&0.245&0.451&1.509&0.127\\
            DETR3D~\cite{wang2022detr3d}&V2-99&900$\times$1600&0.412&0.479&0.641&0.255&0.394&0.845&0.133\\
            BEVDet~\cite{huang2021bevdet}&Swin-B&900$\times$1600&0.424&0.488&0.524&0.242&0.373&0.950&0.148\\
            PETR~\cite{liu2022petr}&V2-99&900$\times$1600&0.441&0.504&0.593&0.249&0.383&0.808&0.132\\
            BEVFormer~\cite{li2022bevformer}&V2-99&900$\times$1600&0.481&0.569&0.582&0.256&0.375&0.378&0.126\\
            BEVDet4D~\cite{huang2022bevdet4d}&Swin-B&640$\times$1600&0.451&0.569&0.511&0.241&0.386&0.301&0.121\\
            PolarFormer~\cite{jiang2022polarformer}&V2-99&900$\times$1600&0.493&0.572&0.556&0.256&0.364&0.440&0.127\\
            PETRv2~\cite{liu2022petrv2}&V2-99&640$\times$1600&0.490&0.582&0.561&0.243&0.361&0.343&\textbf{0.120}\\
            BEVDepth~\cite{li2022bevdepth}&V2-99&640$\times$1600&0.503&0.600&0.445&0.245&0.378&0.320&0.126\\
            BEVStereo~\cite{li2022bevstereo}&V2-99&640$\times$1600&0.525&0.610&0.431&0.246&0.358&0.357&0.138\\
            \hline
            SA-BEV&V2-99&640$\times$1600&\textbf{0.533}&\textbf{0.624}&\textbf{0.430}&\textbf{0.241}&\textbf{0.338}&\textbf{0.282}&0.139\\
            \hline
        \end{tabular}
    \end{center}
\end{table*}

During training, both predictions on the 1/16 and 1/8 scale are supervised. It ensures the first stage head can extract task-specific information and the second stage head can combine the task-specific information with cross-task information. The supervision signals are obtained by projecting point clouds on images following BEVDepth~\cite{li2022bevdepth}. The depth values of the projected points are the depth labels and the points in the 3D boxes are regarded as the foreground. The total loss can be formulated as:
\begin{equation}
    \mathcal{L}=\mathcal{L}_{det} + \frac{\lambda_1}{2}(\mathcal{L}_{S}^{16}+\mathcal{L}_{S}^{8})+\frac{\lambda_2}{2}(\mathcal{L}_{D}^{16}+\mathcal{L}_{D}^{8})
\end{equation}

\section{Experiments}

In this section, we first introduce our experimental settings. Then, comparisons with previous state-of-the-art multi-view 3D detectors are shown. Finally, comprehensive experiments with detailed ablation studies are conducted on SA-BEV to show the effectiveness of each component, i.e. SA-BEVPool, BEV-Paste and MSCT head.

\subsection{Experimental Settings}

\subsubsection{Dataset and Metrics} 
nuScenes~\cite{caesar2020nuscenes} dataset is a large-scale autonomous driving benchmark. It contains 750 scenarios for training, 150 scenarios for validation and 150 scenarios for testing. Each scenario lasts for around 20 seconds and the key samples are annotated at 2Hz. The data collected from six cameras, one LiDAR and five radars are provided to every sample. For 3D object detection, nuScenes Detection Score (NDS) is proposed to capture all aspects of the nuScenes detection tasks. Except mean average precision (mAP), NDS is also related to five types of true positive metrics (TP metrics), including mean Average Translation Error (mATE), mean Average Scale Error (mASE), mean Average Orientation Error (mAOE), mean Average Velocity Error (mAVE), mean Average Attribute Error (mAAE).

\begin{figure*}
    \centering
    \includegraphics[width=17cm]{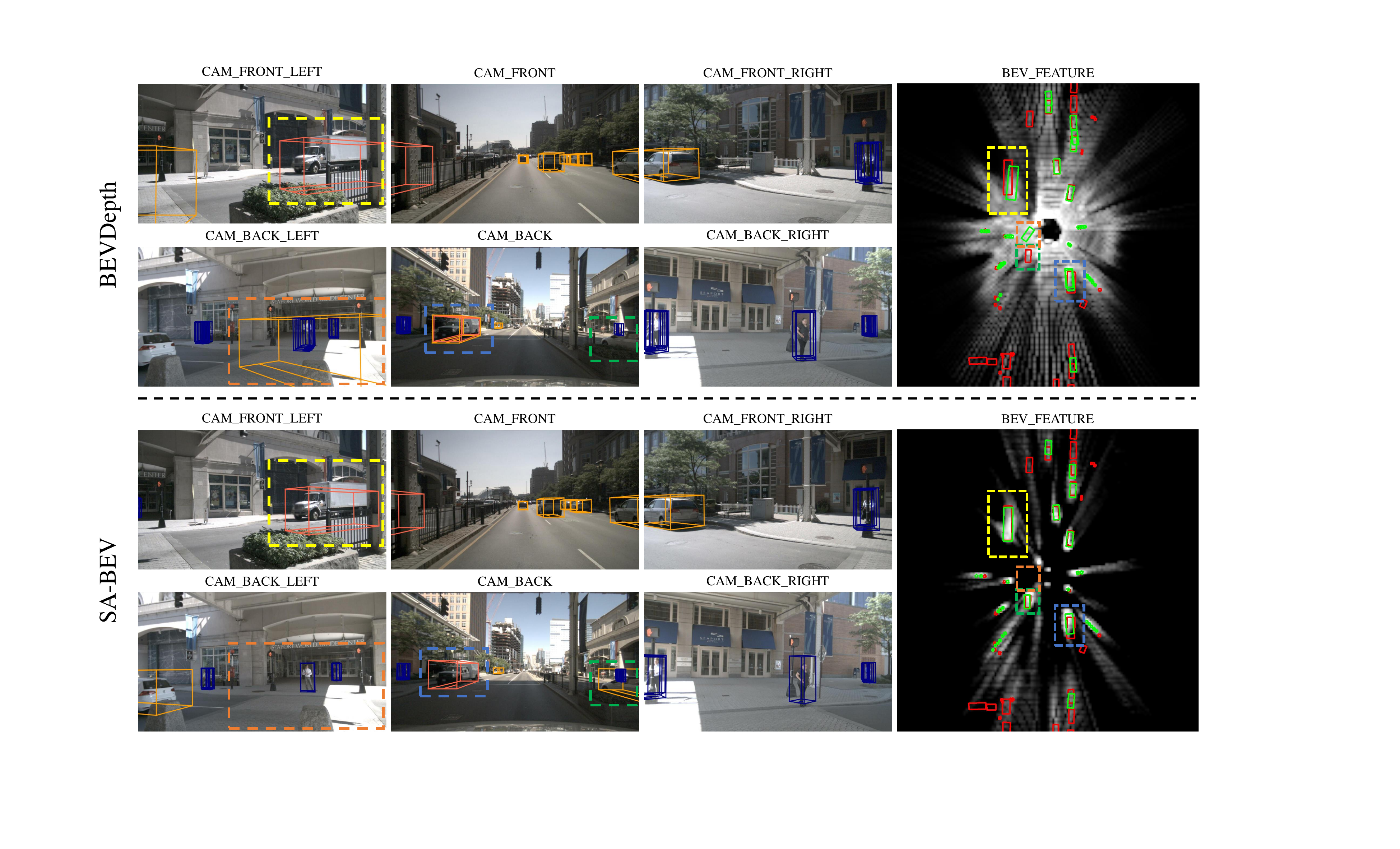}
    \caption{Visualization of detection results on images and BEV features. The red boxes and green boxes on BEV features represent the ground truth and the predicted boxes, respectively. The dashed rectangles illustrate that the prediction of SA-BEV is more precise than BEVDepth.}
    \label{fig:visualization}
\end{figure*}

\begin{table}
    \begin{center}  
        \caption{Comparison with previous state-of-the-art multi-view 3D detectors on the nuScenes \textit{val} set.}
        \label{tab:val}
        \small
        \begin{tabular}{p{2cm}|p{1.7cm}<{\centering}|p{1.3cm}<{\centering}|p{0.6cm}<{\centering}p{0.6cm}<{\centering}}
            \hline
            Method&Backbone&Resolution&mAP$\uparrow$&NDS$\uparrow$\\
            \hline
            FCOS3D~\cite{wang2021fcos3d}&ResNet-101&900$\times$1600&0.343&0.415\\
            DETR3D~\cite{wang2022detr3d}&ResNet-101&900$\times$1600&0.303&0.374\\
            PGD~\cite{wang2022probabilistic}&ResNet-101&900$\times$1600&0.369&0.428\\
            PETR~\cite{liu2022petr}&ResNet-101&512$\times$1408&0.357&0.421\\
            BEVDet~\cite{huang2021bevdet}&Swin-B&900$\times$1600&0.393&0.472\\
            BEVFormer~\cite{li2022bevformer}&ResNet-101&900$\times$1600&0.416&0.517\\
            PETRv2~\cite{liu2022petrv2}&ResNet-101&900$\times$1600&0.421&0.524\\
            BEVDet4D~\cite{huang2022bevdet4d}&Swin-B&640$\times$1600&0.421&0.545\\
            PolarFormer~\cite{jiang2022polarformer}&ResNet-101&900$\times$1600&0.432&0.528\\
            BEVDepth~\cite{li2022bevdepth}&ConvNeXt-B&512$\times$1408&0.462&0.558\\
            BEVStereo~\cite{li2022bevstereo}&ConvNeXt-B&512$\times$1408&0.478&0.575\\
            \hline
            SA-BEV&ConvNeXt-B&512$\times$1408&\textbf{0.479}&\textbf{0.579}\\
            \hline
        \end{tabular}
    \end{center}
\end{table}
\subsubsection{Implementation Detail}

We accomplish our proposed improvements on the network structure of BEVDepth~\cite{li2022bevdepth}. 
Our experiments are implemented based on MMDetection3D with 8 NVIDIA GeForce RTX 3090 GPUs. Models are trained with AdamW~\cite{loshchilovdecoupled} optimizer and gradient clip is utilized. The universal data augmentation we adopt on the image and BEV feature follows the configuration in~\cite{huang2021bevdet}. For the ablation study, we use ResNet-50~\cite{he2016deep} as the image backbone and the image size is downsampled to 256$\times$704. The models are trained for 24 epochs without CBGS strategy~\cite{zhu2019class} for the ablation study. When compared to other methods, the models are trained for 20 epochs with CBGS strategy.

\subsection{Main Results}

\subsubsection{Comparison with State-of-the-Arts}

We compare SA-BEV with state-of-the-art multi-view 3D detectors on nuScenes \textit{test} set and show the results in Table \ref{tab:test}. We take 640$\times$1600 resolution image as input and VoVNet-99~\cite{vovnet} as the image backbone. SA-BEV achieves the best mAP and NDS, 3.0\% and 2.4\% higher than its baseline (i.e. BEVDepth~\cite{li2022bevdepth}). It also exceeds BEVStereo~\cite{li2022bevstereo} by 0.8\% mAP and 1.4\% NDS, which adopts the complicated multi-view stereo structure for more accurate depth estimation. The comparison on nuScenes \textit{val} set is shown in~\ref{tab:val}. It can be found SA-BEV also achieves the best detection precision. The decent results highlight the advantage of the proposed SA-BEV.

\begin{table}
    \begin{center}  
        \caption{Ablation study of component in SA-BEV on the nuScenes \textit{val} set. \textit{pool}, \textit{paste} and \textit{head} denotes SA-BEVPool, BEV-Paste and MSCT head, respectively.}
        \label{tab:component}
        \begin{tabular}{p{2.3cm}|p{0.6cm}<{\centering}p{0.6cm}<{\centering}p{0.6cm}<{\centering}|p{0.8cm}<{\centering}p{0.8cm}<{\centering}}
            \hline
            Baseline&\textit{pool}&\textit{paste}&\textit{head}&mAP$\uparrow$&NDS\\
            \hline
            \multirow{4}{*}{BEVDepth~\cite{li2022bevdepth}}&&&&0.330&0.436\\
            &$\checkmark$&&&0.340&0.449\\
            &$\checkmark$&$\checkmark$&&0.354&0.464\\
            &$\checkmark$&$\checkmark$&$\checkmark$&\textbf{0.365}&\textbf{0.483}\\
            \hline
            \multirow{2}{*}{BEVDet~\cite{huang2021bevdet}}&&&&0.278&0.322\\
            &$\checkmark$&$\checkmark$&&0.304&0.348\\
            \hline
            \multirow{2}{*}{BEVStereo~\cite{li2022bevstereo}}&&&&0.349&0.454\\
            &$\checkmark$&$\checkmark$&&0.364&0.467\\
            \hline
        \end{tabular}
    \end{center}
\end{table}

\subsubsection{Visualization}

We visualize the detection results on images and BEV features in Fig. \ref{fig:visualization}. Compared to BEVDepth, SA-BEV can make more precise predictions with the help of semantic-aware BEV features. For instance, the orange dashed rectangles show that the filtration of the background prevents SA-BEV from making false detection. The yellow dashed rectangles indicate that the semantic-aware BEV feature correctly emphasizes the location of the truck, which results in precise detection. In addition, the green / blue dashed rectangles display that SA-BEV can successfully recall the missed object and remove the redundant detection box.

\subsection{Ablation Study}

\subsubsection{Component Analysis}

We individually evaluate the contributions of SA-BEVPool, BEV-Paste and MSCT head with BEVDepth~\cite{li2022bevdepth} as the baseline. The results are shown in Table \ref{tab:component}. After applying SA-BEVPool, the performance is boosted by 1.0\% and 1.3\% on mAP and NDS. It is further improved by 1.4\% / 1.5\% and 1.1\% / 1.9\% through incorporating BEV-Paste and MSCT head respectively. Finally, we obtain the full model of SA-BEV, which gains 3.5\% and 4.7\% in total, validating its effectiveness. SA-BEVPool and BEV-Paste are also applied to BEVDet~\cite{huang2021bevdet} and BEVStereo~\cite{li2022bevdepth}, increasing 2.6\% / 2.6\% and 1.5\% / 1.3\% respectively on mAP and NDS. It demonstrates that these components can be easily embedded into the existing BEV-based detectors and bring about noticeable precision improvement.

\subsubsection{Semantic-Aware BEV Pooling}
The semantic threshold $T_S$ and semantic threshold $T_D$ in SA-BEVPool control the scale of the valid virtual points. We vary thresholds and show the results in Table \ref{tab:abla_pool}. The results indicate that even a low $T_D$ can sharply reduce the scale of valid virtual points and an appropriate $T_S$ can effectively improve the detection precision. However, a too-high semantic threshold may lead to the loss of foreground information and damage the precision. We set the semantic threshold to 0.25 as a good trade-off. It only needs 1.8\% valid virtual points, resulting in 1\% mAP and 1.3\% NDS performance improvement.

\begin{table}
    \begin{center}  
        \caption{Ablation study of the semantic threshold used in SA-BEVPool. ``Percentage'' denotes the average proportion of valid virtual points.}
        \label{tab:abla_pool}
        \begin{tabular}{p{1cm}<{\centering}p{1cm}<{\centering}|p{1.25cm}<{\centering}p{1.25cm}<{\centering}|p{1.5cm}<{\centering}}
            \hline
            $T_D$&$T_S$&mAP$\uparrow$&NDS$\uparrow$&Percentage\\
            \hline
            -&-&0.330&0.436&100\%\\
            0.0085&-&0.338&0.438&7.92\%\\
            0.0085&0.10&0.339&0.444&3.26\%\\
            0.0085&0.25&\textbf{0.340}&\textbf{0.449}&1.80\%\\
            0.0085&0.50&0.329&0.432&0.89\%\\
            \hline
        \end{tabular}
    \end{center}
\end{table}

\begin{table}
    \begin{center}  
        \caption{Ablation study of BEV-Paste strategy. $N_P$ denotes the average number of frames that are pasted to the original frame.}
        \label{tab:abla_paste}
        \begin{tabular}{p{2.5cm}|p{1.3cm}<{\centering}|p{1.3cm}<{\centering}p{1.3cm}<{\centering}}
            \hline
            Method&$N_P$&mAP$\uparrow$&NDS$\uparrow$\\
            \hline
            \multirow{4}{*}{w/o extra BDA}&0&0.340&0.449\\
            &0.5&0.348&0.453\\
            &1&0.349&0.453\\
            &2&0.349&0.452\\
            \hline
            w/ extra BDA &1&\textbf{0.354}&\textbf{0.464}\\
            \hline
        \end{tabular}
    \end{center}
\end{table}

\begin{table}
    \begin{center}  
        \caption{Ablation study of MSCT head. ``MTD'', ``DS'' and ``MS'' denote the multi-task distillation module, the dual supervision and the utilization of multi-scale image features.}
        \label{tab:abla_head}
        \begin{tabular}{p{1.1cm}<{\centering}p{1.1cm}<{\centering}p{1.1cm}<{\centering}|p{1.2cm}<{\centering}p{1.2cm}<{\centering}}
            \hline
            MTD&DS&MS&mAP$\uparrow$&NDS$\uparrow$\\
            \hline
            &&&0.354&0.464\\
            $\checkmark$&&&0.358&0.468\\
            $\checkmark$&$\checkmark$&&0.361&0.473\\
            &&$\checkmark$&0.361&0.478\\
            $\checkmark$&$\checkmark$&$\checkmark$&\textbf{0.365}&\textbf{0.483}\\
            \hline
        \end{tabular}
    \end{center}
\end{table}

\subsubsection{BEV-Paste}
We conduct experiments with different settings when applying BEV-Paste, including the number of frames pasted to each original frame and whether to utilize extra BDA. The results are shown in Table \ref{tab:abla_paste}. Setting $N_P$ to 0.5 means half of the original frames are augmented by one pasted frame while the others are not augmented. As shown in Table \ref{tab:abla_paste}, the BEV-Paste is not sensitive to the number of pasted frames. Considering that pasting too many frames will cost more time on training detection head because the ground truth objects are increased, setting $N_P$ as 1 is enough. Besides, extra BDA can effectively alleviate data duplication and further improve detection performance. The cooperation of these two points contributes to the performance improvement of 1.4\% mAP and 1.5\% NDS, confirming the effectiveness of BEV-Paste.

\subsubsection{Multi-Scale Cross-Task Head}
The MSCT head contains the Multi-Task Distillation (MTD) module and the Dual Supervision (DS) on prediction from Multi-Scale (MS) image features. A number of experiments are carried out to further verify the effectiveness of each module and the results are shown in Table \ref{tab:abla_head}. Our MTD, DS and MS modules improve NDS performance by 0.4\%, 0.5\% and 1.0\% respectively. We attribute this improvement to the fact that the multi-task distillation module supplements cross-task information, and the dual supervision further promotes the extraction and fusion of task-specific information and cross-task information, as well as the participation of multi-scale image features.

\section{Conclusion and Discussion}

In this paper, we propose SA-BEV to fully utilize the semantic information of images. SA-BEVPool filters out background virtual points and generates semantic-aware BEV features. BEV-Paste then pastes the semantic-aware BEV features of two frames to enhance data diversity. MSCT head introduces multi-task learning and facilitates the optimization of semantic-aware BEV features. 

Our proposed components show strong universality. SA-BEVPool and BEV-Paste can be easily embedded into most BEV-based detectors and bring stable improvements. Besides, we believe that introducing multi-task learning into the generation of semantic-aware BEV features adds a valuable perspective and will inspire future works.

Still, there are limitations in SA-BEV. The thresholds used in SA-BEVPool are manually set, making it hard to achieve optimal performance. BEV-Paste may cause incorrect object overlaps and occlusions when pasting the semantic-aware BEV feature of one frame to another. Those are what we will tackle next. We also would like to extend SA-BEV into a multi-modal detector to activate the complementarity between image and LiDAR. 

\section*{Acknowledgment}
This paper was supported by National Natural Science Foundation of China under grants 62176017 and U20B2069.

{\small
\bibliographystyle{ieee_fullname}
\bibliography{egbib}
}

\newpage
\appendix
\section{More Implementation Details}
\label{sec:a}
\subsection{Data Augmentation}
We augment both images and BEV features following the operation applied in~\cite{huang2021bevdet}. For images, they are first down-sampled to the desired resolution. Then they are processed by random scaling with a range of $[0.94, 1.11]$, random rotating with a range of $[-5.4^{\circ}, 5.4^{\circ}]$ and random flipping with a probability of 0.5. After that, the images are padded and cropped to a uniform shape. For BEV features, augmentation is applied on the virtual points whose features are cumulated to form BEV features. The coordinates of virtual points are processed by random scaling with a range of $[0.95, 1.05]$, random flipping of the X and Y axes with a probability of 0.5 and random rotating with a range of $[-22.5^{\circ}, 22.5^{\circ}]$. Augmenting virtual points rather than BEV features themselves can generate more accurate augmented BEV features because the bilinear sampling is not required by the former. The additional BEV data augmentation (BDA) used by BEV-Paste also follows the above settings.

\subsection{Detection Configuration}

We use the detection head of CenterPoint~\cite{yin2021center} to detect 3D objects from semantic-aware BEV features and follow the settings used in BEVDepth~\cite{li2022bevdepth}. The LiDAR coordinate system of nuScenes is used to represent the coordinate of points in the BEV space. The X and Y coordinates are in the range of $[-51.2, 51.2]$, and the Z coordinate is in the range of $[-5, 3]$. The BEV space is divided into pillars for cumulating virtual point features. When the resolution of input images is $256\times704$, the pillars are in the size of $[0.8, 0.8, 8]$ and the BEV features are in the shape of $128\times128$. For larger input images, the pillars are in the size of $[0.4, 0.4, 8]$ and the BEV features are in the shape of $256\times256$.

\section{More Experiment Results}
\label{sec:b}
We change the image backbone of SA-BEV to ResNet-101 when processing $512\times1408$ resolution images and compare it with other methods that also utilize ResNet-101 as their backbone. The results are shown in Table~\ref{tab:r101}. SA-BEV achieves the best mAP and NDS, 2.9\% and 1.4\% higher than its baseline (i.e. BEVDepth~\cite{li2022bevdepth}). It also exceeds other start-of-the-art methods that take $900\times1600$ resolution images as input. This comparison further proves the effectiveness of SA-BEV.
\begin{table}
    \begin{center}  
        \caption{Comparison with previous state-of-the-art multi-view 3D detectors on the nuScenes \textit{val} set.}
        \label{tab:r101}
        \small
        \begin{tabular}{l|c|c|p{0.55cm}<{\centering}p{0.55cm}<{\centering}}
            \hline
            Method&Backbone&Resolution&mAP$\uparrow$&NDS$\uparrow$\\
            \hline
            FCOS3D~\cite{wang2021fcos3d}&ResNet-101&900$\times$1600&0.343&0.415\\
            DETR3D~\cite{wang2022detr3d}&ResNet-101&900$\times$1600&0.303&0.374\\
            PGD~\cite{wang2022probabilistic}&ResNet-101&900$\times$1600&0.369&0.428\\
            PETR~\cite{liu2022petr}&ResNet-101&512$\times$1408&0.357&0.421\\
            BEVFormer~\cite{li2022bevformer}&ResNet-101&900$\times$1600&0.416&0.517\\
            PETRv2~\cite{liu2022petrv2}&ResNet-101&900$\times$1600&0.421&0.524\\
            PolarFormer~\cite{jiang2022polarformer}&ResNet-101&900$\times$1600&0.432&0.528\\
            BEVDepth~\cite{li2022bevdepth}&ResNet-101&512$\times$1408&0.412&0.535\\
            \hline
            SA-BEV&ResNet-101&512$\times$1408&\textbf{0.441}&\textbf{0.549}\\
            \hline
        \end{tabular}
    \end{center}
\end{table}

\begin{figure}
    \centering
    \includegraphics[width=8cm]{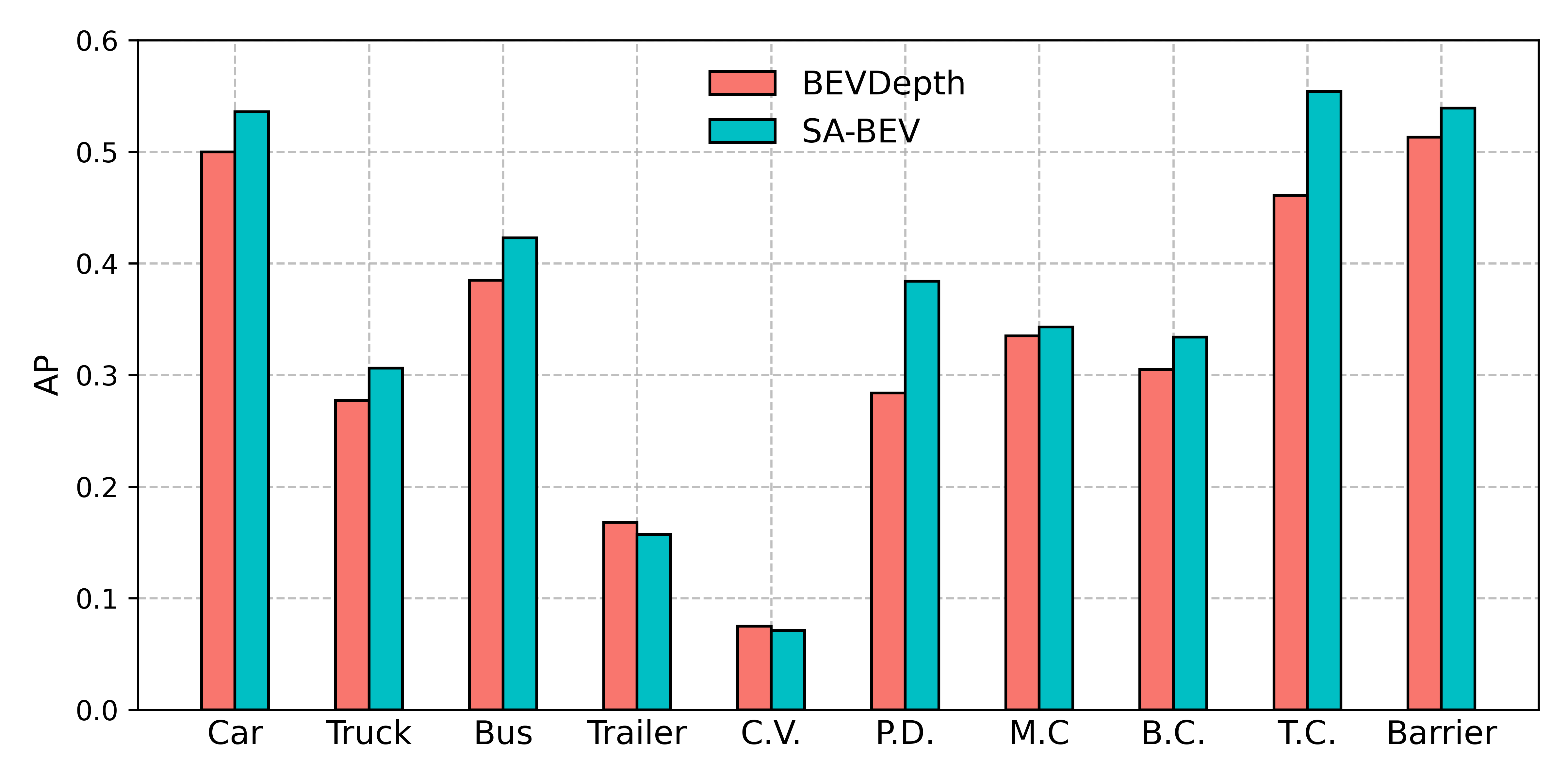}
    \caption{Comparison of BEVDepth and SA-BEV on AP for each category. C.V., P.D, M.C., B.C. and T.C. are the abbreviations of construction vehicle, pedestrian, motorcycle, bicycle and traffic cone respectively.}
    \label{fig:class_ap}
\end{figure}

\begin{figure*}
    \centering
    \includegraphics[width=16cm]{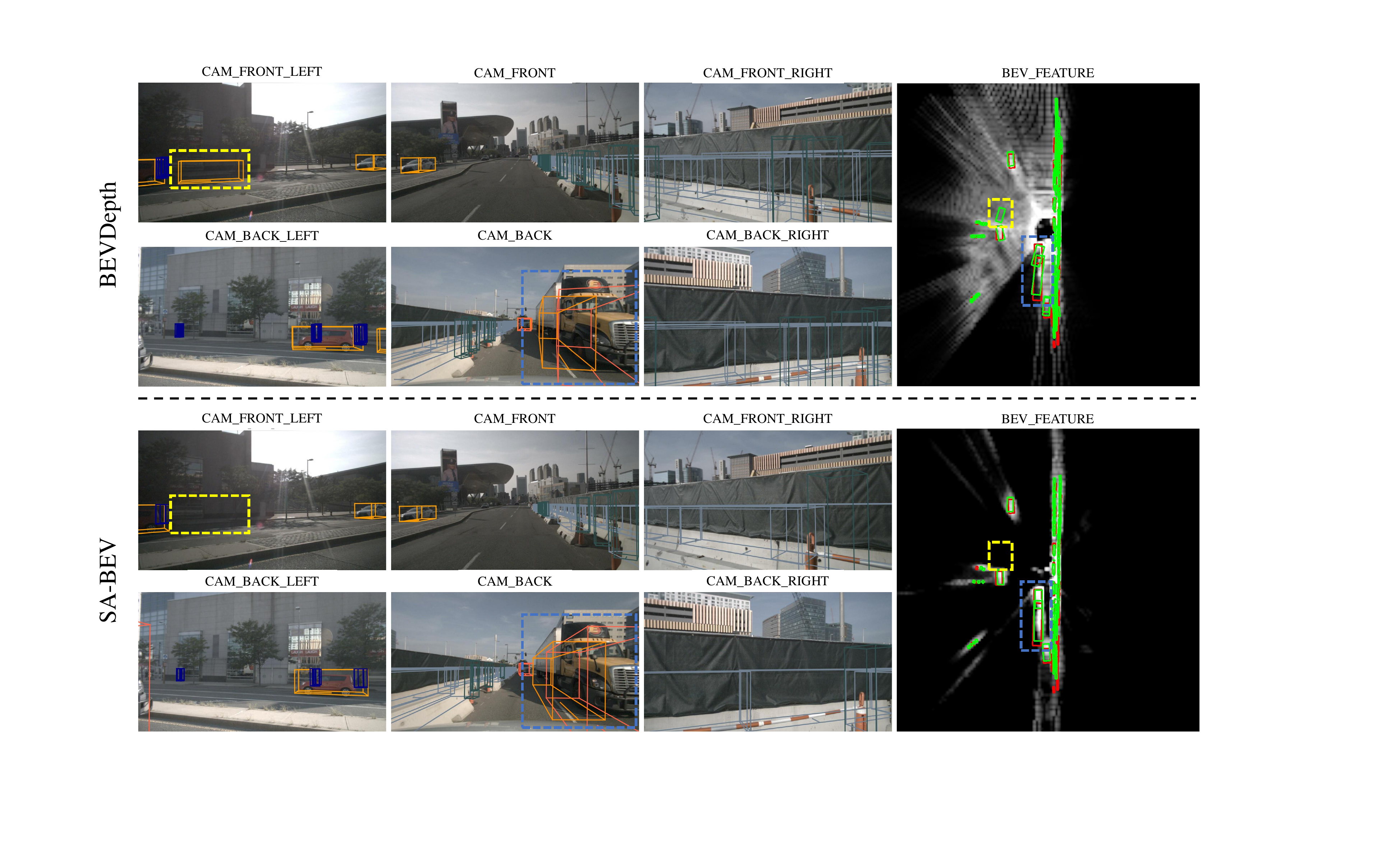}
    \includegraphics[width=16cm]{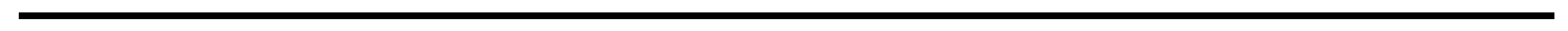}
    \includegraphics[width=16cm]{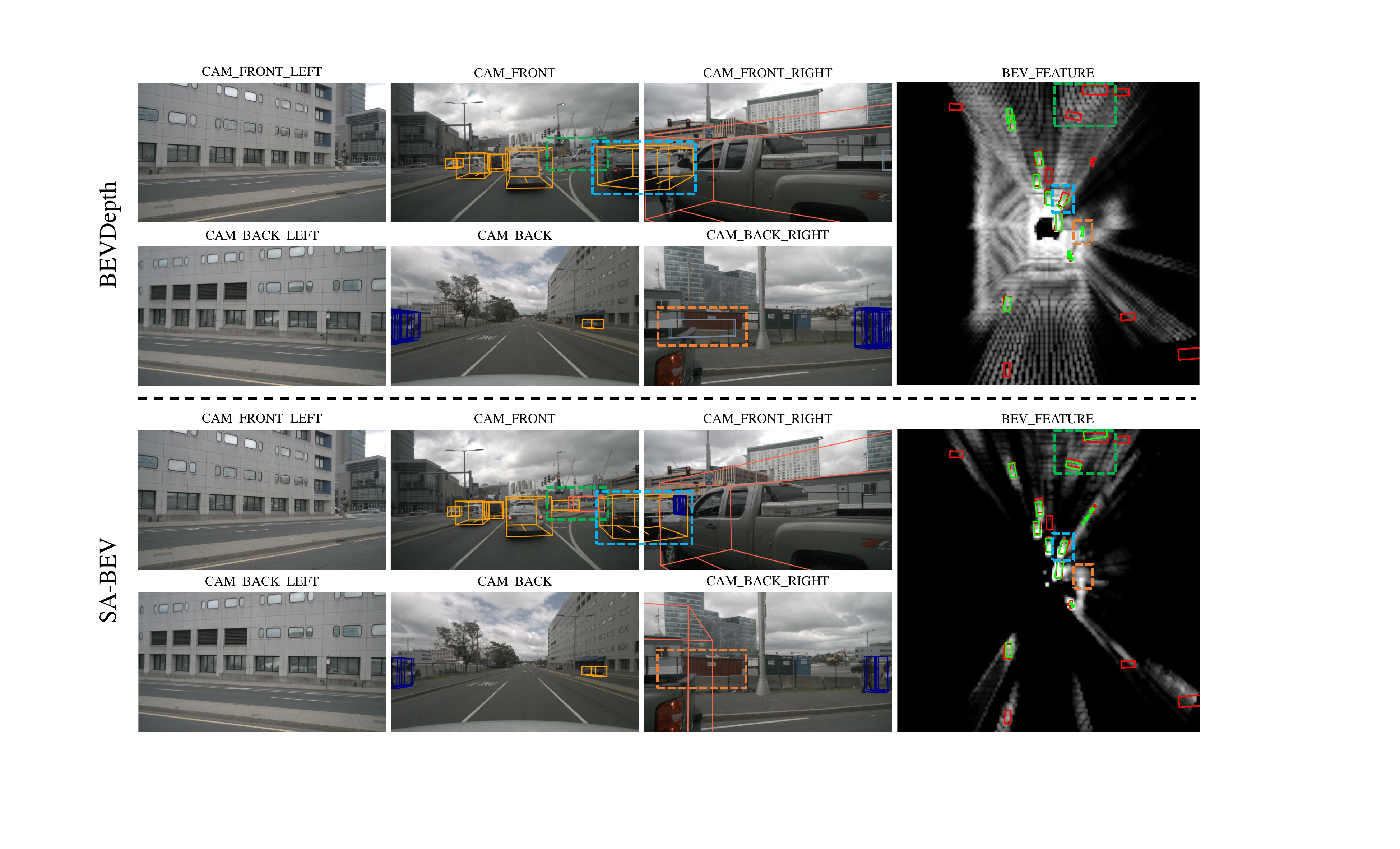}
    \caption{Visualization results on images and BEV features. The red boxes and green boxes on BEV features represent the ground truth and the predicted boxes, respectively. The dashed rectangles illustrate that the prediction of SA-BEV is more precise than BEVDepth.}
    \label{fig:vis}
\end{figure*}

We also compare the detection precision of BEVDepth and SA-BEV in each category and show the results in Fig.~\ref{fig:class_ap}. SA-BEV achieves better precision than BEVDepth in most of the categories. For instance, the APs on pedestrian and traffic cone are increased by about 10\%, and the APs on car, truck, bus and bicycle are increased by about 3\%. The greater improvement in pedestrian and traffic cone categories indicates that the semantic-aware BEV features effectively preserve the information of small scale objects that is more likely to be submerged by the large proportion of background information.

\section{More Visualization Results}
\label{sec:c}

We provide more visualization results of BEVDepth and SA-BEV in Fig.~\ref{fig:vis}. With the help of semantic-aware BEV features, SA-BEV can recall objects in the far distance and identify the false truth precisely. Besides, SA-BEV generally predicts more accurate locations and directions of the objects, which is also important in actual practice.

\end{document}